\documentclass[runningheads]{llncs}
\usepackage{graphicx}

\usepackage{tikz}
\usepackage{comment}
\usepackage{amsmath,amssymb} 
\usepackage{color}
\usepackage{amssymb}
\usepackage{booktabs}
\usepackage{enumitem}
\usepackage{verbatim}
\usepackage{xcolor}

\usepackage{subcaption}

\newcommand\blfootnote[1]{%
  \begingroup
  \renewcommand\thefootnote{}\footnote{#1}%
  \addtocounter{footnote}{-1}%
  \endgroup
}

\begin{document}
\pagestyle{headings}
\mainmatter

\title{Video Question Answering with Iterative Video-Text Co-Tokenization} %

\titlerunning{Video QA CoTokenization}
%
\author{AJ Piergiovanni$^\dagger$ \and Kairo Morton$^{*\dagger}$ \and Weicheng Kuo$^\dagger$ \and Michael S. Ryoo$^\dagger$ \and Anelia Angelova$^\dagger$}
\authorrunning{A. Piergiovanni et al.}
%
\institute{$^\dagger$Google Research, $^*$MIT\\
\email{ajpiergi@google.com}}
\maketitle

\vspace{-1cm}
\begin{abstract}
Video question answering is a challenging task that requires understanding jointly 
the language input, the visual information in individual video frames, as well as the temporal information about the events occurring in the video. In this paper, we propose a novel multi-stream video encoder for video question answering that uses multiple video inputs and a new video-text iterative co-tokenization approach to answer a variety of questions related to videos. We experimentally evaluate the model on several datasets, such as MSRVTT-QA, MSVD-QA, IVQA, outperforming the previous state-of-the-art by large margins. Simultaneously, our model reduces the required GFLOPs from 150-360 to only 67, producing a highly efficient video question answering model. \blfootnote{*Work done as an intern at Google}\footnote{Code: \url{https://sites.google.com/view/videoqa-cotokenization}}
\keywords{video question answering, video-text joint learning, video understanding, efficient vision models}
\end{abstract}

\section{Introduction}

\begin{figure}[t]
   \centering
   \includegraphics[width=0.8\linewidth]{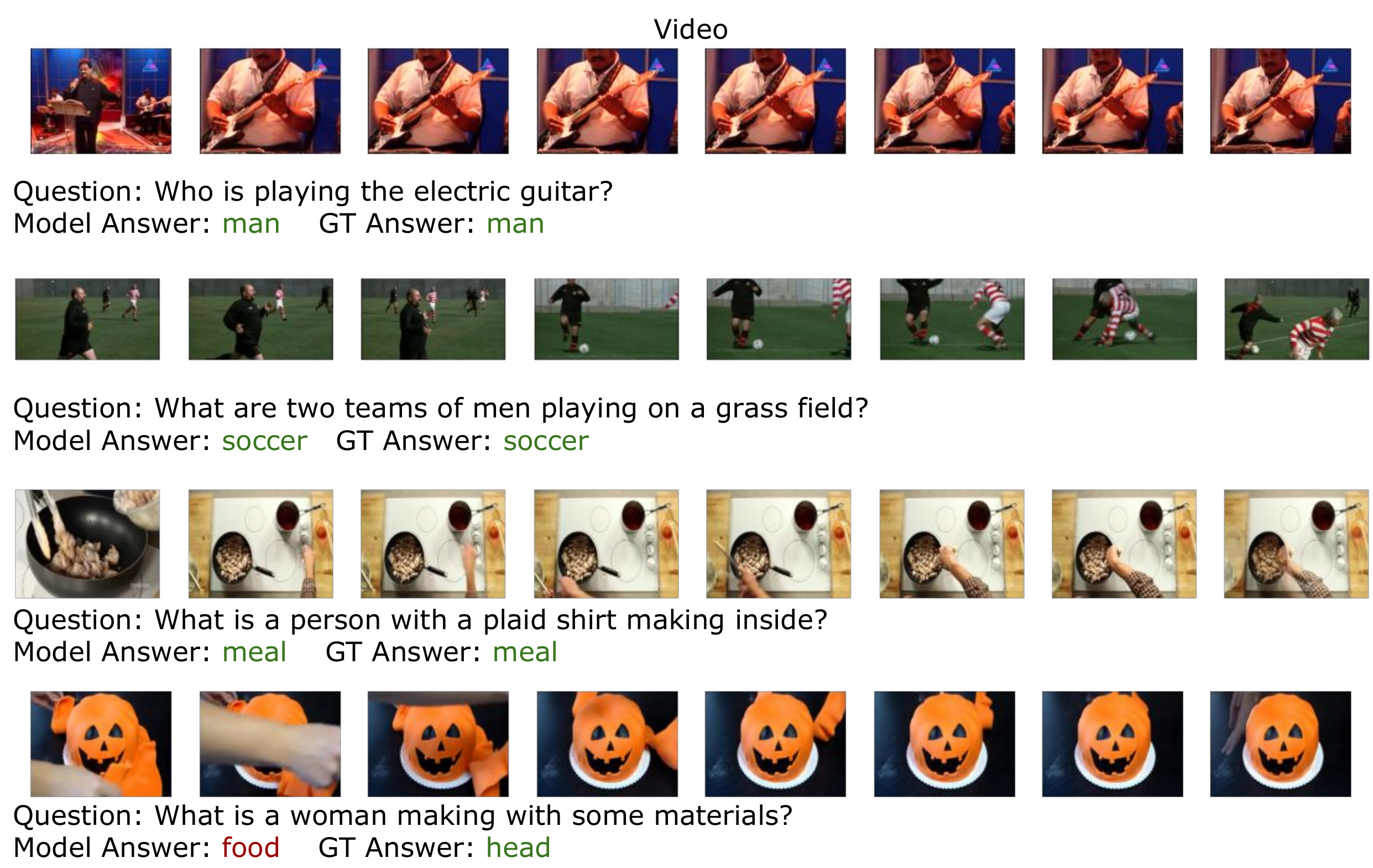}
  \caption{We consider the challenging task of Video Question Answering which is a multi-modal information-seeking task, where natural language questions or tasks specifications are issued towards a video. The answers are in natural language open-vocabulary text. Example VideoQA outputs of our model are shown.}
   \label{fig:teaser}
 \end{figure}

Video Question and Answering (VideoQA)~\cite{jang2017tgifqa,zhu2018uncovering,li2019beyondrnns,kim2020modality,xu2017videoqa} targets the challenging problem of 
answering a variety of questions about a video, which requires both natural language understanding and understanding of the video content, including reasoning about activities, objects, sequence of events, and interactions within the video. 
VideoQA is the video counterpart of 
Visual Question and Answering (VQA)~\cite{das2017visual,jiang2020in,agrawal2015vqa,goyal2017making}, a long-standing task in computer vision of answering questions towards an image.
%
VideoQA is a very important multi-modal visual-language task for natural interaction with videos,
aiming to understand what is happening in a video with the help of text-based specification (Figure~\ref{fig:teaser}). It can satisfy information needs from videos and allow for rich user engagement such as searching for highlights, events, objects, or specific scenes in a video, e.g., ``Where is the first goal scored in the game?'', ``How many goals were scored?'' ``Why was the umpire ruling considered controversial?''.

VideoQA has the inherent challenges of VQA tasks: it needs to understand the visual and language inputs and how they relate to each other. 
Additionally, VideoQA, needs to address multiple challenging video understanding tasks, such as action recognition, action detection and segmentation~\cite{Rohrbach2012adatabase,carreira2017quo,lea2017temporal}, but unlike them, needs to work in the open-set domain\footnote{Unlike standard video understanding tasks, e.g., action recognition, action segmentation, where the set of action classes is pre-defined and known, the open-set VideoQA involves answering natural questions about novel or unseen actions and/or objects.}, where questions can be issued about unseen object categories or unknown activities. 
VideoQA needs a deeper understanding of the video input to begin with, which requires adequate spatio-temporal understanding.

VideoQA faces other challenges too: it requires the ability to process larger visual inputs e.g., 30-100x the number of frames than VQA, and to answer much harder questions, for example, why a certain event is happening, or what has happened before a certain action, where the timeframes might span small or large portions of the video. 
Furthermore, VideoQA inherits the efficiency challenges of video processing, which very few of the prior approaches have addressed.

The visual question and answering problem has been explored by many in the image and video research domains~\cite{agrawal2015vqa,howtovqa69m,alamri2019avsd,lei2018tvqa}. However, 
many previous works with regards to this problem 
utilize multiple backbone feature extractors for individual image frames, for the video input and for the text input separately and apply cross-attention between modalities as an after thought following feature extraction \cite{li2020hero,howtovqa69m}. However, we believe, and will show in this paper, that the interaction between these modalities during the visual feature extraction process is key in attempting to solve this problem. 
We propose to jointly learn the video and language representations, where their interaction allows reasoning across these two modalities. 
Furthermore, by viewing question text not just as additional information or as a part of the task, but as the lens through which visual data is understood, it is clear that these cross-modality interactions must occur at various stages throughout the process of extracting visual features.
Specifically, we propose a novel video-text iterative co-tokenization approach which learns efficient joint  representations iteratively\footnote{We consider video inputs of 32 frames which spans up to 10 seconds of video.}.

Our work effectively demonstrates the benefits of using text during the video understanding tasks for a variety of open ended VideoQA tasks and shows improvements over the state-of-the-art (SOTA) on MSVD-QA~\cite{xu2017msvd-qa},
MSRVTT-QA~\cite{xu2016msrvtt}, and IVQA \cite{howtovqa69m} datasets.  Extensive ablation experiments confirm the benefit of each component.

Our paper makes the following contributions: 
\begin{itemize}
    \item Novel video-language interaction learning for videos and text, especially focusing on learning both spatial and temporal features, which outperforms several SOTA on several VideoQA datasets.
    \item Novel multi-stream video encoder with iterative video-text co-tokenization which uses multiple inputs and iteratively selects efficient features 
    \item An efficient approach, greatly reducing the FLOPs over baselines, which is important to save compute for video methods and allows scalability.
\end{itemize}

\section{Related work}

\textbf{Video understanding.}
Video understanding is a fundamental visual recognition task, conducted in video inputs~\cite{3dconv,tran2014c3d,simonyan2014two,carreira2017quo,xie2018rethinking,zhou2018temporal,tran2018closer,feichtenhofer2018slowfast,wu2021towards,timeception,bertasius2021timesformer,ryoo2019assemblenet,ryoo2021tokenlearner_neurips,korbar2019scsampler,eco2018,lin2019tsm}. What makes it challenging is the joint processing of spatial and temporal information and the sheer volume and diversity of visual inputs. Video understanding tasks include action classification, action detection, object segmentation in videos, etc. Text is not used in the training and recognition process.

\textbf{Video and language.}
Using language alongside videos~\cite{lei2021clipbert,chen2019weakly} has opened several interesting problems in multi-modal video-language learning and language-guided video understanding, for example, VideoQA~\cite{jang2017tgifqa,zhu2018uncovering}, video captioning~\cite{huang2020multimodal,krishna2017dense,deng2021sketch,zhou2018e2e}, text-to-video retrieval~\cite{xu2016mstvtt,rohrbach2015adataset,lei2021clipbert,zhou2018towards,bain2021frozenintime,dong2021dual,multiquery}, 
referring expression comprehension for videos~\cite{bellver2020refvos,hendricks2017localizing,khoreva2018video} and others~\cite{hendricks2017localizing,gao2017tall,wang2020temporally}.
In addition to the challenges of the video understanding tasks, video+language tasks bring in their own challenges of understanding text in the context of video, analyzing the video content according to the text input, or in some tasks, natural language text generation.

Previous video-language methods use pre-training from separate video models and text models which are typically pre-trained on disjoint datasets~\cite{sun2019videobert,gabeur2020multi,zhu2020actbert,li2020hero,miech2019Howto100m,lin2021vx2text,yu2021learning}. Pre-training video and text jointly has been shown to be very beneficial to a number of downstream tasks~\cite{miech2020e2e,zhu2020actbert}. 
End-to-end joint training with multi-modal inputs from the target datasets is also gaining popularity recently~\cite{lei2021clipbert}.
In the above-mentioned approaches, Transformer-based models~\cite{attention}, adapted to videos are often used to join the two modalities~\cite{sun2019videobert,zhu2020actbert}, e.g., with masked-learning or other  objectives~\cite{zhou2018e2e,vilbert2020}; in other works, standard representations e.g., I3D, S3D~\cite{carreira2017quo,xie2018rethinking} for video and word2vec embeddings for text, 
have also been explored for joint training~\cite{miech2020e2e}.

\textbf{VideoQA.}
Video Question and Answering covers a broad range of questions towards a video which require both understanding of the video input and the input text~\cite{jang2017tgifqa,zhu2018uncovering,li2019beyondrnns,kim2020modality,xu2017videoqa,alamri2019avsd,maharaj2017vidqa,alamri2019avsd,xu2021trafficqa,garcia2020knowit,zadeh2019social,yu2019activitynet,zhukov2019crosstalk,xiao2021nextqa,Wang2019vatex,lei2018tvqa,tapaswi2016movieqa,li2021value,howtovqa69m}.
Several research methods have been developed for VideoQA~\cite{rohrbach2017movie,kim2017deepstory,zhu2018uncovering,hori2019older,minh2020hier,yu2018ajoint,fan2019hetero,kim2018multimodal,tsai2019video,park2021bridge,jiang2020reasoning,li2019beyondrnns,le2020neuralreasoning,xue2018abetter,chadha2021iperceive}. 
Large-data pre-training is commonly used for VideoQA~\cite{zhu2020actbert,miech2020e2e}. 
ClipBert~\cite{lei2021clipbert} propose end-to-end learning of video and language tasks, which are typically tested on text-to-video retrieval and VideoQA~\cite{jang2017tgifqa}.
Kim et al.~\cite{kim2020modality} propose multi-modal question and answering where additionally a text input such as captioning can be included in the input.

\textbf{Video-language cross-modal learning.} Image-language cross-modal learning has been developed for cross-modality retrieval, e.g., initially for images and text~\cite{gong2014improving,donahue2015long} and for videos and text~\cite{xu2015jointly,dong2019dual}. These approaches are followed by using the popular transformer architecture~\cite{attention} to encode separately each modality into a joint space, applied to both image-text~\cite{miech2021thinking} and to video-text retrieval~\cite{sun2019videobert,miech2019Howto100m,miech2021thinking}.  
A similar transformer-based architecture is commonly used for learning jointly image and language features via joint attention or cross-modal co-attention modules~\cite{vilbert2020,chen2020uniter,tan2019lxmert,su2020vlbert,li2020oscar,yu2019deepmodular} 
This is also common in video-language joint learning, e.g.,~\cite{zhu2020actbert,lei2021clipbert}. 
In the context of VideoQA,
Zhu et al.~\cite{zhu2020actbert} use the transformer architecture with pre-computed bounding boxes from an object detector, the features of which are 
fed to transformer layers.   
Hero~\cite{li2020hero} propose cross-modal transformer for multi-modal fusion based on Uniter~\cite{chen2020uniter}. 
Fan et al.~\cite{fan2019hetero} also propose a multimodal fusion layer.
In complement to these cross-modal approaches, ours specifically learns efficient representations to reflect the interactions of these modalities at each level of abstraction of feature learning. 
Our multistream formulation is also conceptually related to SlowFast \cite{feichtenhofer2018slowfast} for videos and multi-stream object detection networks for object understanding, where different network streams were utilized to look at different resolutions.

 \begin{figure}[t]
   \centering
  \includegraphics[width=0.5\linewidth]{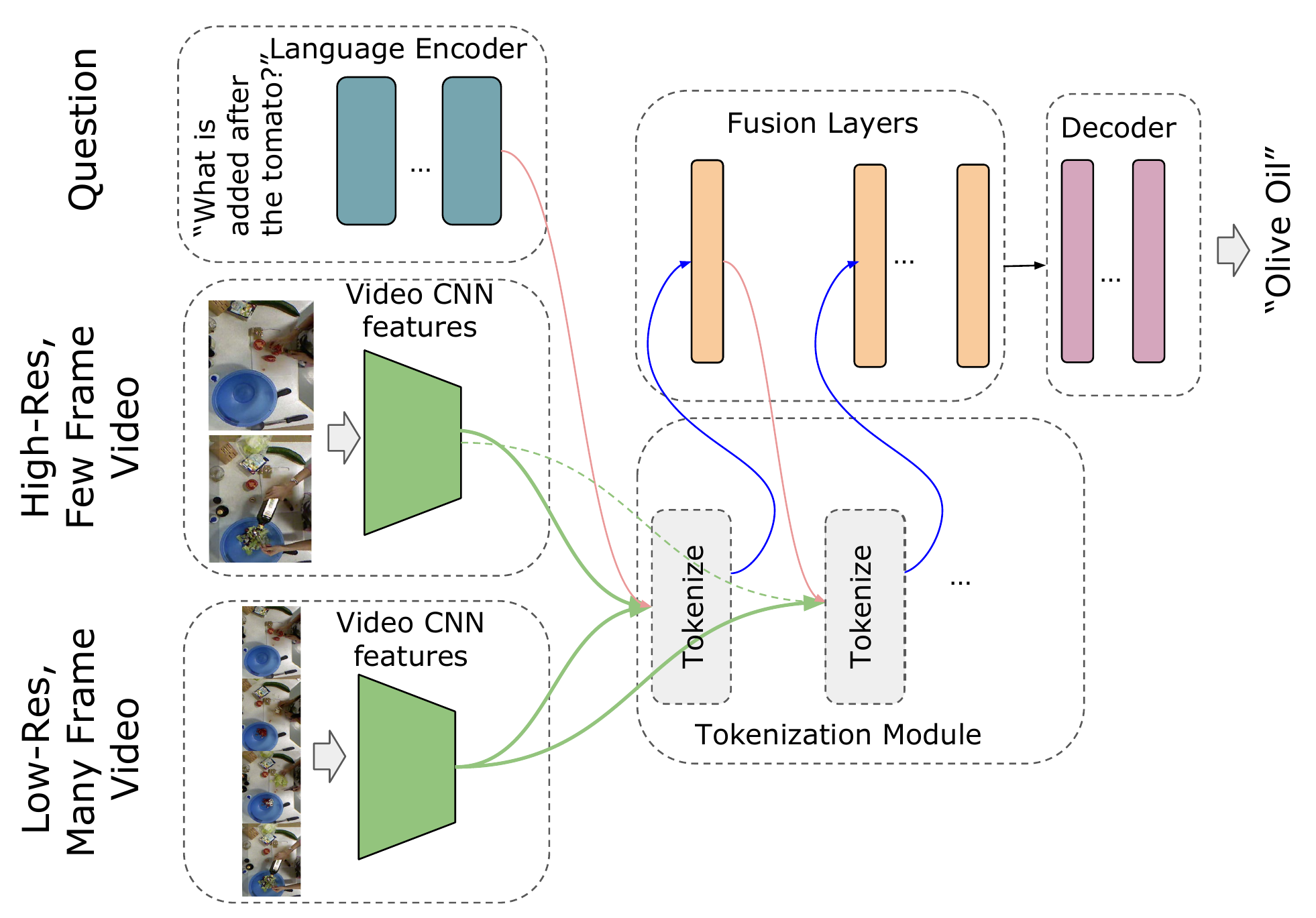}
  \caption{Main architecture. Video features interact with text features. This is done efficiently using multi-stream video encoder to learn spatial and temporal features fusion with text features.  Iterative co-tokenization fusion module learns a reduced number of useful tokens by progressively using the fused video-language representation to affect the next tokenization (red arrows). This results in a very efficient approach as compact tokenized representations are learned throughout. 
  The decoder, which produces the answer, is a standard text-generation decoder.}
   \label{fig:main}
   \vspace{-0.5cm}
 \end{figure}
 
 \begin{figure}[t]
   \centering
   \includegraphics[width=0.8\linewidth]{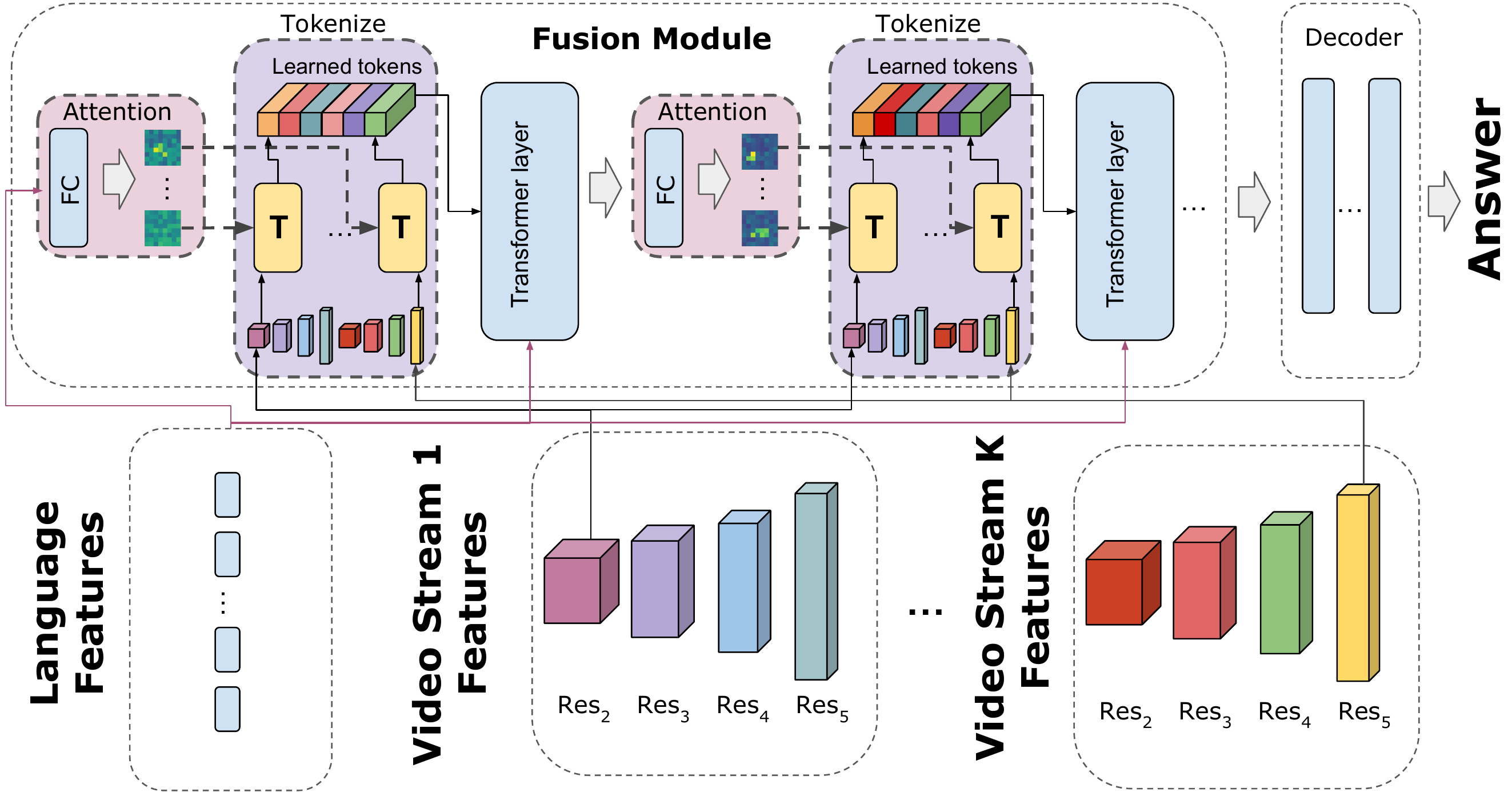}
   \caption{Illustration of the iterative fusion module in detail. The features are used to generate $N$ attention maps, which are applied to each stream and (optionally) scale of the video features. This results in a fixed number of tokens for the $K$ streams and $S$ scales ($K$ and $S$ are typically very small). The tokens are fed through a transformer layer, generating a new feature, which is used to generate new attention maps. This processes is repeated multiple times. The yellow boxes (denoted as `T') represent the application of an attention map to a feature map.}
   \vspace{-0.6cm}
   \label{fig:fusion}
 \end{figure}

\section{Approach}

\subsection{Multi-Stream Video Encoder Overview}

Our approach to the challenging VideoQA task, is video-language learning where early visual and language features are jointly learned.   

Specifically, we propose a multi-stream video encoder that is capable of efficiently learning both spatial and temporal features, and a new fusion method that can adaptively combine the video and text features.
Our multi-stream encoder features the following components: First, the video features are extracted forming multiple learnable video representation inputs (Section~\ref{sec:timeframes}); the text input is also preliminarily encoded. Secondly, a condensed representation for each input stream is learned via learned tokenization (Section~\ref{sec:tl}). Importantly, the features are combined by an iterative video-text co-tokenization fusion mechanism which learns the most appropriate compact feature representations iteratively based on the previous features (Section~\ref{sec:joining}).
Figure~\ref{fig:main} gives an overview of the main architecture.

At a high-level, our approach can be thought of as an encoder-decoder structure, however here the multi-stream video encoder is tasked with interdependent, video-language feature learning, which produces compact and efficient features. 
A text generation decoder directly outputs natural language free-form text.

\vspace{-0.1cm}
\subsection{Video understanding at different timeframes}
\label{sec:timeframes}
Actions and events in videos span a wide range of timeframes. Questions about videos also require different time scales. For example, `What color is the apple?' only requires understanding a single frame, while `What happens after cutting the apple?' requires localizing two specific actions segments in the video in order to answer, and `What is being made in the video?' requires understanding the entire video.
Being able to answer all these questions, with a model that is computationally efficient is a challenge, as, unlike images, the model must consider features of large video inputs with multiple spatial and temporal resolutions.

To address this, we propose a multi-stream approach with fusion, that extracts video features at various time and space scales. We consider them jointly together, as well as with the text/question input.

\textbf{Video features sub-inputs.} The video encoder, $V$, takes a video as input, e.g., a $x=T\times H\times W$ tensor where $T$ is the number of frames and $H$ and $W$ are height/width of the image frame. It processes the video, producing the output features $f_v$. The baseline model uses a single video input, while the multi-stream model takes the video at different space- and time-scales as input. For example, one stream can take many frames at low spatial resolution, while another stream can take few frames at high spatial resolution. We can here use any number of video streams, each learning different spatio-temporal features.
Importantly, subsequent components of the multi-stream visual encoder will enable learning various inter-relations of these inputs. This is critical for VideoQA, since the question can refer to the full video or specific spatial or temporal segments in it, or require the comprehension of events or actions in the video across different duration timeframes, or be very specifically pinpointed in time and space.

The features from the multi-stream visual encoder can further be mutli-scale, taking features from different points in the network, e.g., after each residual block in X3D. We denote the output(s) of the multi-stream encoder as $f_v = V(x)$, where $f_v$ in general a set of multi-scale features over multiple feature levels: $f_v = \{f_{vi}\}_{i=1}^S$ where $S$ is the total number of video features from different streams.

\vspace{-0.1cm}
\subsection{Learning to tokenization}
\label{sec:tl}
One important aspect for fusing multi-modal models is to achieve effective representations of each modality but also do so efficiently. A possible naive approach is to apply either global average pooling, reducing the entire vision stream to a single representation, or concatenate many frames together. The first averages all the spatio-temporal information into a single representation, while the latter maintains a lot of redundant information and is computationally expensive.

Instead, we here first `compress' each modality by learning to tokenize, based on TokenLearner~\cite{ryoo2021tokenlearner_neurips}. The advantage is that it learns to select a small set of tokens, conditioned on the inputs. This greatly reduces the number of spatio-temporal tokens, while learning to maintain the needed information. Due to its adaptive nature, it can change the tokens based on the video and text inputs, allowing it to focus on the most important information.

We modify the TokenLearner approach to better suit the problem here, since in~\cite{ryoo2021tokenlearner_neurips} it is only applied to single image frames. In our case, we also have both videos and text as our inputs, and videos are of multiple resolutions. We here enable both video and text to condition how video tokens are learned. Intuitively, this will allow TokenLearner to better select important spatio-temporal regions in the video not only based on visual information itself but also based on the text, and adaptively tokenize such regions.

Specifically, we learn to tokenize by first taking the input representation $r$ with shape $L \times F$, where $L$ is the length of the sequence and $F$ is the feature dimension.  This initially is the text feature, $r=f_t$. We also take a specific video feature $f_{vi}$, where $i$ is one of the multi-scale, multi-stream features which has shape $T\times H\times W\times C$.

Given $r$, we learn $\phi(r)$ (implemented with a linear layer in our version), to produce a $L \times (T \cdot H \cdot W)$ tensor where $T \cdot H \cdot W$ is the temporal and spatial size of the video feature. Another linear layer is used to make the feature have shape $C \times (T \cdot H \cdot W)$.  This is reshaped and added with the video feature: $f=\phi(r) + f_{vi}$. 
Next, the function $\psi(\cdot)$ (implemented with convolutional layers) is applied to convert $f$ into a feature with shape $T\times H \times W \times N$, where $N$ is the number of desired tokens. 
A softmax function ($\sigma$) is applied over this, along the $N$-axis, selecting the spatio-temporal features for each token. 
Enabling the spatial attention mechanism, this is multiplied with the video representation, $f_{vi}$. 
More specifically, the attention mask $\sigma(\psi(\phi(r) + f_{vi}))$ is transposed to shape $N\times T\times H \times W$, and is tensor-dot-producted with $f_{vi}$ while treating $T\times H \times W$ as the dimensions to contract.
Overall, this can be expressed as:
\begin{equation}
\label{eq:TL}
f_0^i=\sigma(\psi(\phi(r) + f_{vi})) \cdot f_{vi}
\end{equation}
The resulting feature representation $f_0^i$ will have a shape of $N\times C$, abstracting the entire video as a set of $N$ tokens per each video feature $f_{vi}$. 

This allows the tokenizations to reduce many video streams into a few tokens.
Generally, for our multi-stream, multi-scale inputs, the final feature representation $f_0$ is obtained by concatenating all $f_0^i$ in the first axis, forming a representation with shape of $(NS) \times C$ where $S$ is the number of scales and streams that are used.
Importantly, these adaptive tokens are learned according to the optimization loss of the final task, which aims to improve accuracy of the produced outputs or answers. These tokens are then fused with a learning mechanism described in the below subsections.

\subsection{Video-text iterative co-tokenization}
\label{sec:joining}
%
We here describe how we apply the above-mentioned joint tokenization approach iteratively at various levels of feature abstraction. 
We utilize self-attention transformer layers to combine the text and video features. Different from previous works, we will use the features of the transformer layers to produce a few informative tokens from the data (blue arrows on Fig.~\ref{fig:main}), which are fed to the next transformer fusion layer (red arrows on Fig.~\ref{fig:main}). Importantly, we use the token learning mechanism presented above at every layer; This allows the model to change its (video feature) selections differently at different layers.

To do this, we start by using tokenization to select $NS$ visual tokens from the video input, as described above, initially using the text feature. These are then concatenated with the encoded text representation: $[f_t, f_0]$ along the token axis. These are then passed through 1 transformer layer ($H$), $r_1=H([f_t, f_0])$. The outputs, $r_1$ are then used as input to Eq. \ref{eq:TL} to generate $NS$ new tokens, $f_{1}$, which are again concatenated with the text representation, added to the previous encoded, and passed through the next transformer layer: $r_l = H([f_t, f_{l-1}]+r_{l-1})$ for $l > 1$. The tokenziation is done for each of the multi-stream, multi-scale features $f_{vi}$, and is repeated $L$ times, where $L$ is the number of transformer layers (Fig.~\ref{fig:fusion}). 

This approach allows the model to adaptively and iteratively select different visual features, from multiple scales and streams, refining the input to best align with the text. It results in a highly efficient method, due to the iterative tokenization.

\vspace{-0.3cm}
\subsection{Implementation details}
\vspace{-0.1cm}
For better comparison to SOTA, we use standard model components, the T5 language model~\cite{T5}, and for video, we use popular video representations: 3D ResNets, 2D ResNets + temporal pooling, and X3D \cite{feichtenhofer2020x3d}. The encoded text and video features (learnable end-to-end) are entered into the multi-stream video encoder for learning the interaction between these features. 
A language-based decoder, again T5~\cite{T5}, then takes the fused features to generate the output text. We set $N=8, S=4, C=768; K=3$ 

\textbf{Video Preprocessing:}
To preprocess the video data, each individual frame is resized a fixed size, e.g., 224 by 224 pixels, and normalized such that the value for each pixel ranges from -1 to 1. We sample $T$ frames from each video, evenly spaced across the video, i.e., the frames-per-second per video varies to maximize the temporal extent. In the multi-stream setting, $T$ and $H\times W$ are different for each stream. We describe these settings for the models below.

\textbf{Text Preprocessing:}
The text is tokenized using T5's standard 32k word vocabulary, with a max length of 32 tokens per example for both input and output. The model is trained to minimize the output per-token cross-entropy. 

\begin{table}[]
      \centering
\caption{MSRVTT-QA~\cite{xu2016msrvtt} and MSVD-QA~\cite{xu2017msvd-qa} datasets. Accuracy (\%).}
\label{tab:msr}
    \scalebox{1.0}{  
    \begin{tabular}{l|c c c}
    \toprule
     Model  &MSRVTT-QA &MSVD-QA &GFLOPs \\
    \midrule
ST-VQA~\cite{jang2017tgifqa}  &31.3 &30.9 &-\\
Co-Mem~\cite{gao2018comem} &31.7 &32.0 &-\\
AMU~\cite{xu2017videoqa} &32.0 &32.5 &-\\
HME~\cite{fan2019hetero} &33.7 &33.0 &-\\
HRA~\cite{Chowdhury2018hier} &34.4 &35.1 &-\\
HCRN~\cite{minh2020hier} &36.1 &35.6 &-\\
ClipBERT~\cite{lei2021clipbert} 8x2  &37.4 &- &-\\
OCRL+LOGNet~\cite{dang2021object} &38.2 &36.0 &-\\
\midrule
VQA-T~\cite{howtovqa69m} with  HowTo100M  &40.4 &43.5 &\textcolor{gray}{75+}\\
VQA-T~\cite{howtovqa69m} with HowToVQA69M  &41.5 &46.3 &\textcolor{gray}{75+}\\
SiaSamRea \cite{yu2021learning} & 41.6 & 45.5 & -\\
MERLOT\cite{merlot} with YT-Temporal-180M*~\cite{merlot}  &43.1 &- & -\\
\midrule 
Ours with HowTo100M       &\textbf{45.7} &\textbf{48.6} &67\\
    \bottomrule
    \end{tabular}
    }
\end{table}

\begin{table}
\centering
\vspace{-0.2cm}
\caption{Results comparing to SOTA approaches on the IVQA~\cite{howtovqa69m} dataset.}
\label{tab:ivqa}
    \scalebox{1.0}{  
    \begin{tabular}{l|c c c}
    \toprule
     Model  &Pre-training dataset &Accuracy (\%) & GFLOPs\\
    \midrule

VQA-T~\cite{howtovqa69m} &HowTo100M  &28.1 &-\\
VQA-T~\cite{howtovqa69m} &HowToVQA69M  &35.4 &-\\
\midrule 
Ours              &HowTo100M &\textbf{38.2} &67 \\
    \bottomrule
    \end{tabular}
 }
 \vspace{-0.5cm}
\end{table}
\vspace{-3mm}
\section{Experiments}
\vspace{-3mm}
We conduct experiments across different VideoQA datasets in order to determine the benefits of the approach, pretraining, efficiency and scaling.

\vspace{-4mm}
\subsection{Datasets}
\textbf{IVQA} The IVQA dataset \cite{howtovqa69m} is a new, human annotated dataset for VideoQA consisting of `how-to' videos. It has 10k clips with 1 question and 5 answers per question. It follows an evaluation metric similar to the VQA2.0 dataset \cite{agrawal2015vqa}, where its accuracy is computed for 5 choose 4 ground-truth (GT) answers.

\textbf{MSRVTT-QA} \cite{xu2017videoqa} is based on the MSRVTT descriptions dataset \cite{xu2016msrvtt}, with automatically generated QA pairs from the descriptions. It has 243k VideoQA pairs and is evaluated by answer accuracy.

\textbf{MSVD-QA} \cite{xu2017videoqa} is based on the MSVD datasets with automatically generated QA pairs. It has 50k VideoQA pairs and is evaluated by answer accuracy.

\textbf{TGIF-QA}~\cite{jang2017tgifqa} consists of short GIF video clips with accompanying questions and answers related to the video. There are four types of questions ranging from questions about objects in still frames, to activities, repetition counting and sequences of events. TGIF-QA is evaluated in the multiple-choice answer setting, which we show is trivial for a strong (and language-only) model (see supp.). Instead, we use TGIF-QA in an open-vocabulary generative setting to study the effects of our approach on both single-frame and temporal-based questions.

\subsection{Evaluation setup}
We follow the standard metrics for each dataset, described above. Since our model is generative, we use three different evaluation settings. (1) Open-ended generation, where the text decoder is used as-is with beam search~\cite{beamsearch}, and generates any text. We then check string equality of the generated text to the GT answer. (2) Vocabulary-specific fully-connected (FC) layer, where we use the target vocabulary for each dataset from \cite{howtovqa69m}, and train a new FC layer on top of the final language features to classify the answer. (3) Masked-vocabulary generation, 
where we keep the text decoder from the model, but mask out tokens not in the target vocabulary. This lets us preserve the learned token embedding, but restrict the vocabulary to that of previous works, so as to be directly comparable.


\section{Experimental Results}

We compare our model with the SOTA approaches (Section~\ref{sec:sota}) and then examine the model in a number of ablations. Section~\ref{sec:efficient}, reports on model efficiency.

\vspace{-5mm}
\subsection{VideoQA results. Comparison to SOTA}
\label{sec:sota}

Table~\ref{tab:msr} shows our results comparing to the MSRVTT-QA and MSVD-QA datasets, which are commonly used for VideoQA evaluation. 
As seen, our results outperform the SOTA, even though our pretraining is done on the weaker captioning dataset (HowTo100M), instead of the VideoQA counterpart, HowToVQA69M, which has been shown to be superior for VideoQA tasks~\cite{howtovqa69m}.

Table~\ref{tab:ivqa} compares our approach on the challenging new IVQA dataset. As seen, our results outperform SOTA with both HowToVQA69M and HowTo100M. Importantly, with the same pre-training HowTo100M, our approach outperforms the SOTA by more than 10\% in absolute values.

In Table 1 of the supp. material, we show a surprising but instructive result on TGIF-QA, where our method, using a medium size pretrained text model T5 (T5-Base)~\cite{T5}, is able to accomplish close to 100\% results on the multiple choice questions. This is due to the fact that the limited selection of answers is easy to guess even without video (it performs randomly for the action counting category as it needs the video input for these). We acknowledge that this contemporary text model is stronger than text models used previously, but is important to note that the multiple-choice setting is too easy. We experiment with the harder open-vocabulary setting in our ablations. The datasets evaluated above, MSRVTT-QA, MSVD-QA, IVQA are open-ended and thus do not suffer from this problem.

\begin{figure}[t]
   \centering
   \includegraphics[width=0.8\linewidth]{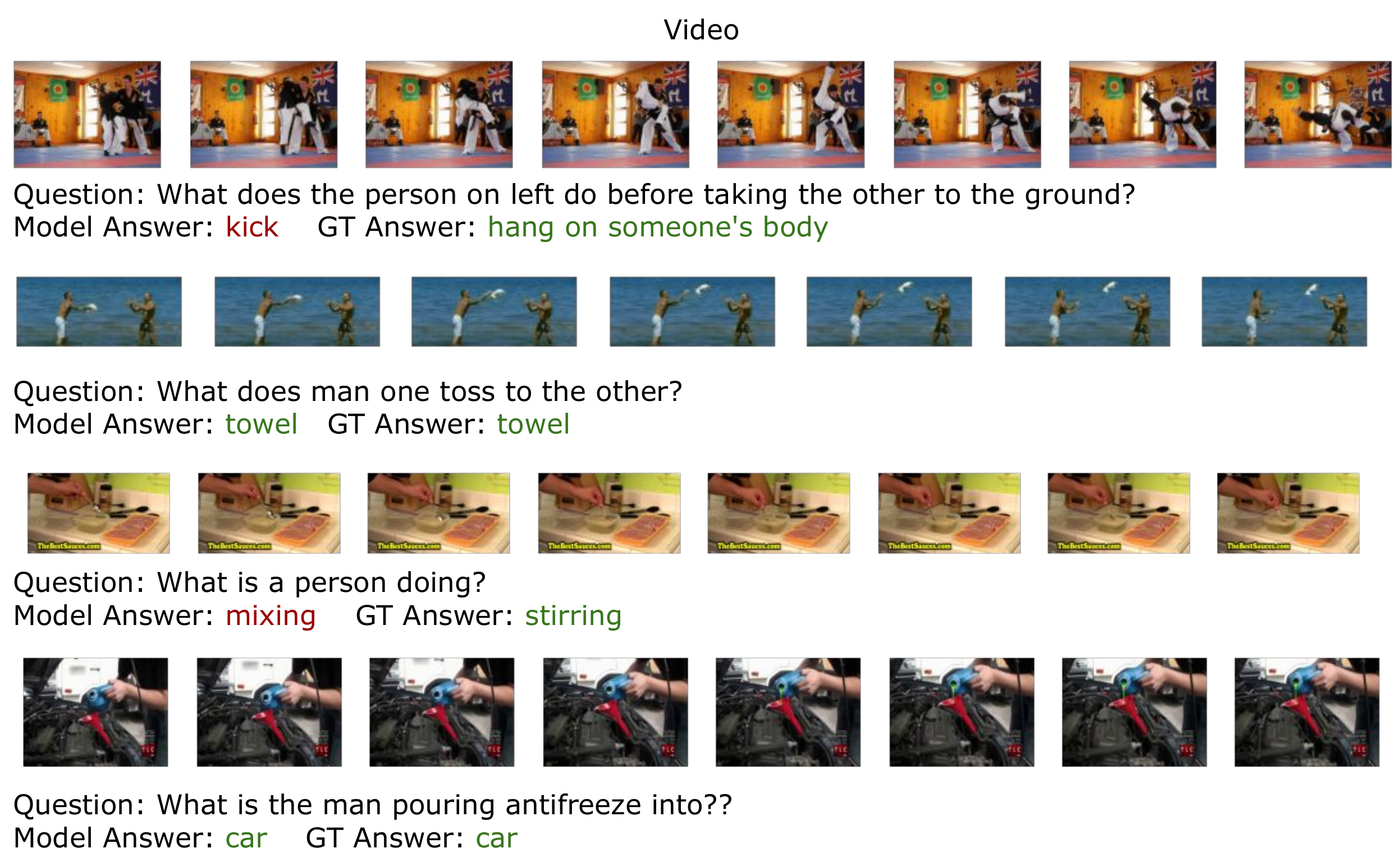}
  \caption{Example results of our method.}
   \label{fig:results}
   \vspace{-6mm}
 \end{figure}

\begin{figure}
    \centering
    \includegraphics[width=0.9\linewidth]{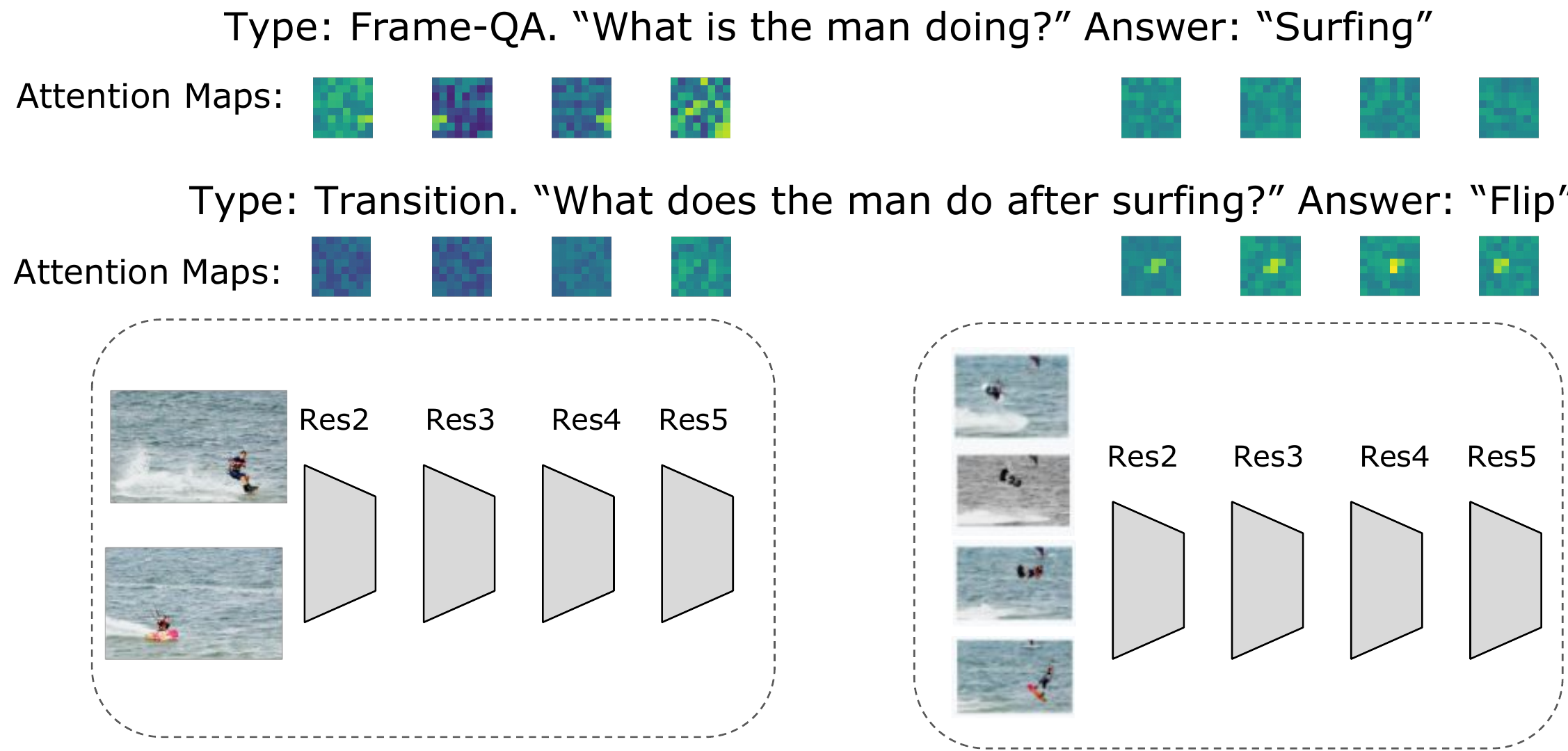}
    \caption{Visualization of the learned attentions for two different types of questions. We can see it learns to select the spatial stream for the ``Frame-QA'' type of question, where temporal information is not needed, but for the ``Transition'' question, it focuses more heavily on the stream with many frames.}
    \label{fig:connection}
\end{figure}

\begin{figure}
    \centering
    \includegraphics[width=0.75\linewidth]{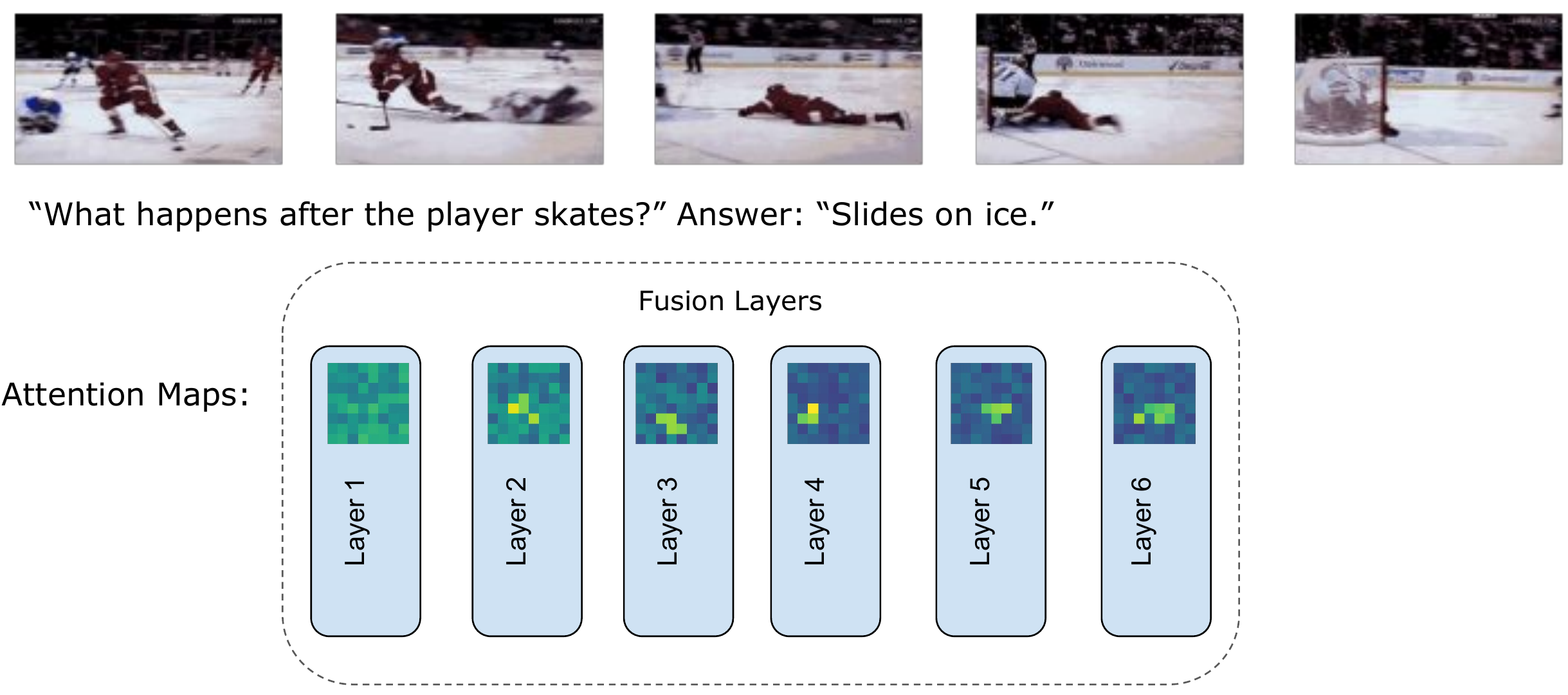}
    \caption{Attention maps changing after each layer they are applied to. The first one focuses over the whole video, each after focuses on more specific regions. In this example, it captures the hockey player falling and sliding on the ice.}
    \label{fig:iterative}
\end{figure}

\textbf{Visualizations.} In Fig.~\ref{fig:results} we show example results of the approach. In Fig. \ref{fig:connection}, we visualize the learned attention maps for different types of questions. In Fig. \ref{fig:iterative}, we see a set of attention maps for each transformer layer. These figures show that the model is adapting the tokens based on both the video and text.

\vspace{-0.2cm}
\subsection{Open-vocabulary answer generation}
In the previous SOTA results (as presented in Section~\ref{sec:sota}), prior approaches still use a limited vocabulary of answers, e.g., 4000 answers \cite{howtovqa69m}. 
However in real-life scenarios, it is desirable that the generated answers are free-form text. In this section we show that generating open-vocabulary answers is a much more challenging setting for VideoQA. With our experiments, we would like to encourage future results to report this setting, as well.

In Table~\ref{tab:vocab-settings} we compare the open-ended, vs. FC vocabulary vs. masked vocabulary. The open vocabulary is the most challenging, as it requires the model to generate (in free form) the exact sequence of tokens to match the GT word. The FC vocabulary is restricted to 4k tokens, matching \cite{howtovqa69m} vocabulary, but this model throws away the learned token embeddings, losing information. The masked vocabulary preserves all the token information, but removes the unneeded tokens, closely matching the previous settings, while maintaining the learned features.

\begin{table}[]
    \centering
    \caption{Comparing the performance of the same model with a fixed vocabulary (as reported in SOTA) and with open vocabulary, which is more challenging.}
    \label{tab:vocab-settings}
    \begin{tabular}{l|ccc}
    \toprule
    Vocabulary / Dataset & IVQA &	MSRVTT-QA &	MSVD-QA \\
    \midrule
    Open 32k vocabulary & 21.4 & 33.7 & 32.5 \\
    Fixed 4k vocabulary (FC) & 37.4 & 42.9 & 45.9\\
    Fixed 4k vocabulary (Masked) & \textbf{38.2} & \textbf{45.7} & \textbf{48.6} \\
    \bottomrule
    \end{tabular}
    \vspace{-0.3cm}
\end{table}

\vspace{-0.3cm}
\subsection{Fusion techniques}
\label{sec:fusion}
We conduct ablations on the multi-stream video encoder itself and its fusion mechanisms. Table~\ref{tab:fusion-exps} has the results -- open vocabulary and no pre-training are used in these experiments. We see that each of the components of multi-stream video encoder contribute, where we note large collective contributions of the novel tokenization-based fusion techniques and also improvements compared to a single-stream model. Our approach is also very efficient, despite being multi-stream, due to the iterative tokenization approach and efficient backbones.
Note that multiple frame models hurt the single-frame QA setting, but greatly benefit the time-based questions. Further, we see the most gains from the approach in the multi-frame questions, showing the benefit of the approach. 

\vspace{-0.1cm}
\subsection{Video multi-stream encoder model efficiency}
\label{sec:efficient}
In Table~\ref{tab:scaling} we experiment with the effect of scaling the model to more streams and larger models and see that our approach is much more efficient, in addition to being accurate, even with more than one streams.
Our multi-stream approach, with only 67 GFLOPs, allows using various model sizes to improve performance and save FLOPs/params.
We also note that to our estimates, the popular video vision transformer-based model ViViT~\cite{vivit} requires 2010 GFLOPs if adapted to VideoQA with T5, which would be infeasible both for training or inference (ViViT does not report VideoQA results).  We used 64 TPUs for 72 hours to pretrain (4608 TPU hours) and 4 TPUs for 8 hours to finetune (32 TPU hours). Overall, this is fairly modest, e.g., compared to MERLOT \cite{merlot} which used 1024 TPUs for 30 hours (30k TPU hours) for pretraining.

\vspace{-0.6cm}
\begin{table}[]
    \centering
        \caption{Ablations on multi-stream video encoder and fusion techniques. These models are trained from scratch and in the open-vocab generative setting. The ablations are cumulative, e.g., the last row uses 2-stream + Tok + MS + Co-Tok. Adding each component benefits. Tokenization reduces FLOPs notably and brings small performance improvements, whereas multi-scale and iterative co-tokenization bring larger improvements for very modest additional FLOPs. These models were trained with 32 frames, 224x224 images. The two streams are 32x224x224 and 32x128x128.}
    \label{tab:fusion-exps}
     \scalebox{0.9}{  
    \begin{tabular}{l|c|ccc|c|c}
    \toprule
    Model & GFLOPs & TGIF & TGIF Action & TGIF Trans. & IVQA & MSRVTT \\
    & & \begin{tabular}{@{}c@{}}Frame-QA \\ Single frame\end{tabular} & \begin{tabular}{@{}c@{}}(What happens \\ X times?)\end{tabular} & \begin{tabular}{@{}c@{}}(What happens \\ after X?)\end{tabular} & & QA\\
    \midrule
    Single-frame  & -- & 24.4 & 	0.7 & 	1.5	&	8.4&	7.2 \\
    Single-stream & 150 & 21.5 & 8.2 &	9.2	&	14.2&	24.8\\
    \midrule
2-stream   & 47 & 24.2 &	8.8 &	9.2	&	14.5&	24.7\\
     + Transformer & 49 & 24.5 &	9.1 &	10.9 &		14.4&	25.3\\
     + Tokenization & 40 & 24.7 &	9.7	& 11.6	&	14.9&	25.5 \\
     + Multi-Scale  & 41 & 26.2 &	11.5 &	12.2	&	15.2&	26.2\\
     + Iterative Co-Tok. & 42 & \textbf{27.3} &	\textbf{11.8} &	\textbf{12.5}	&	\textbf{15.5} &	\textbf{27.6}\\
    \bottomrule
    \end{tabular}
    }
\end{table}

\vspace{-1cm}
\begin{table}[]
    \centering
       \caption{
    Efficiency comparisons. While we use multiple streams, they take much fewer FLOPs. They also outperform the strongest X3D-XL model. Open vocabulary setting, no pre-training. 2-str (X3D-S 8x224x224, X3D-M 16x112x112),
3-str (X3D-S 8x224x224, X3D-M 16x112x112, X3D-M 32x64x64),
3-str (3x X3D-L 8x224x224, 16x112x112, 32x64x64). Note that these models use fewer frames than the ones in Table~\ref{tab:fusion-exps}.
    }
    \label{tab:scaling}
    \scriptsize
    \begin{tabular}{l|ccc|cc}
    \toprule
    Model & IVQA &	MSRVTT-QA&	MSVD-QA	&	GFLOPS & Params \\
    \midrule
X3D-S & 9.4 &	24.8 &	22.4& 82 & 311M\\
X3D-XL & 10.3 &	27.8 &	23.2 & 150 & 380M \\
2D-RN-50 & 2.2 &	6.4	 & 6.5& 306 &	332M\\
3D-RN-50 & 8.9 &	24.4 & 23.2	&	362 &	341M\\
\midrule
2-stream (X3D-S, X3D-M) & 9.2 &	25.3 &	23.5 & 40	 & 321M \\
3-stream (X3D-S, X3D-M, X3D-M) & 10.3	& 28.2 & 23.8	&	42&	335M\\
3-stream (3x X3D-L) & \textbf{12.4} & \textbf{30.5} & \textbf{25.7}  & 67 & 345M\\
\bottomrule
    \end{tabular}
    \vspace{-0.5cm}
\end{table}

\vspace{-0.3cm}
\begin{table}[]
    \centering
     \caption{Comparing different pretrianing methods for the VideoQA task. All use the single-stream baseline X3D-M model.}
    \label{tab:pretrainin}
    \begin{tabular}{l|ccc}
    \toprule
    Method & IVQA &	MSRVTT-QA &	MSVD-QA \\
    \midrule
    Random init & 7.2 & 8.3 & 4.8 \\
     Kinetics-600 + T5 & 10.8 & 26.8 & 23.2 \\
     Contrastive (YouTube8M)   & 10.8 & 25.8 & 26.2  \\
     Contrastive (HowTo100M)   & 11.0 &	26.2 & 25.5 \\
      Generative (YouTube8M) & 15.2 & 27.8 & 28.2 \\
     Generative (HowTo100M)    & \textbf{15.6} & \textbf{29.4} & \textbf{28.6} \\
    \bottomrule
    \end{tabular}
\end{table}
\vspace{-0.2cm}

\subsection{Pretraining}
In Table~\ref{tab:pretrainin} we explore different versions of pretraining (PT): contrastive, generative, Kinetics~\cite{kay2017kinetics} classification, etc. for the VideoQA task. The models are pretrained using either HowTo100M with the automatic captions, YouTube8M~\cite{YouTube-8M} with automatic captions (when YouTube8M is split into 5 second clips, similar to HowTO100M clip duration, it has 250M clips, about 2x the size of HowTo100M). We pretrain the models using contrastive training (e.g., \cite{miech2019Howto100m}) or language-generative training, in a completion setting where half the caption is used as input and half used as the target text. We also compare to Kinetics 600 classification pretraining of the video model and using the pretrained T5 model~\cite{T5} for text.
We find HowTo100M to be strongest of these pretrainings, and for our model, the generative training was better than contrastive. Both were generally better than independent Kinetics + T5 pretraining.

\subsection{Exploring the effects of temporal features}
In Table~\ref{tab:temporal-features} we compare the effects of temporal features on TGIF-QA and MSRVTT-QA, IVQA. We compare the ResNet 1-frame, 16-frame and 32-frame model, as well as the X3D video models. We can see that, for single frame questions, the 2D ResNet does better. However, for temporal questions, X3D is far superior. This further confirms the benefit of multi-stream models for VideoQA tasks. 

\begin{table}[]
    \centering
    \vspace{-0.5cm}
     \caption{Comparing temporal features. Here the ResNet is pretrained on image-text data, e.g., CC12M. We find that single-frame PT ResNet does very well on the single frame QAs, but quite poorly on the temporal questions. Adding frames help, but in general, we find that learning video specific features helps the most. 2D ResNets are shown in the top half and X3D-M in the bottom. }
    \label{tab:temporal-features}
    \begin{tabular}{l|ccc|c|c}
    \toprule
    Model & TGIF Frame-QA & TGIF Action & TGIF Transition & IVQA & MSRVTT-QA \\
    &  (Single frame) & \begin{tabular}{@{}c@{}}(What happens \\ X times?)\end{tabular} & \begin{tabular}{@{}c@{}}(What happens \\ after X?)\end{tabular} & &\\
    \midrule
    1 frame &	37.4 &	0.8 &	2.9 &	12.7 &	22.3 \\
    16 frame &	35.4 &	2.5 &	3.5	&	18.6 &	23.5 \\
    32 frame &	36.2 &	3.3 &	3.9	&	19.3 &	23.9 \\
    \midrule
    16 frame & 26.2 & 12.4 & 13.4 & 13.5 &	24.5 \\
    32 frame &	25.5 & 13.1	& 15.4 & 14.2 &	24.8 \\
    \bottomrule
    \end{tabular}
\end{table}
\vspace{-0.5cm}

\section{Conclusions}
In this paper we propose a novel multi-stream video text encoder with iterative video-text co-tokenization, which efficiently extracts and fuses information from video and text inputs.
We demonstrate its benefits to challenging VideoQA tasks, outperforming SOTA on the standard benchmarks and metrics and also report results in the more challenging open-vocabulary setting. Our approach is efficient taking only 67 GFLOPs. One limitation of the approach is for longer answer. Though our model is able to generate longer answers as it is generative, the challenge is the evaluation metric. The metric we use is string equality for accuracy. However, for long answers this will be quite hard.


\clearpage
%
%
\bibliographystyle{splncs04}
\bibliography{egbib}

\clearpage
\appendix

\section{Additional experiments}

Table~\ref{tab:tgif} shows the results on the TGIF-QA dataset~\cite{jang2017tgifqa}.
It is an interesting result, where our method, using a medium size pretrained text model T5 (T5-Base)~\cite{T5}, is able to accomplish close to 100\% accuracy on the multiple-choice questions which are for the `Actions'  and `Transition' types of questions. This is due to the fact that the limited selection of answers is easy to guess even without video\footnote{The Text-only method performs randomly for the action counting category as it definitely needs the video input to correctly answer how many times an action is performed.}. The questions from the `FrameQA' type are not multiple choice and thus are harder to guess. We acknowledge that this contemporary text model is stronger than the previous text models, but is important to note that these two tasks likely had been mostly language understanding ones. We also note that the datasets evaluated in the main paper, MSRVTT-QA, MSVD-QA, IVQA are much more challenging and do not suffer from this problem. 
We also compare the results of using our approach with the open vocabulary output (last row of Table~\ref{tab:tgif}).
Note that when using the open vocabulary text-generative setting  on TGIF-QA, the tasks are significantly harder, even when using both video and text, with much accuracy lower numbers showing the difficulty of this setting.

\begin{table*}[]
    \begin{center}
\caption{TGIF-QA. We note that when using a contemporary pre-trained text model (T5-Base), our results achieve almost 100\% on the multiple-choice questions, alluding the answers are easy to guess.}
\label{tab:tgif}
    \scalebox{1.0}{  
    \begin{tabular}{l|ccc}
    \toprule
     Model &Action &Transition &FrameQA \\ 
    \midrule
ST-VQA(R+C)    &60.8 &67.1 &49.3 \\
Co-Memory(R+F) &68.2 &74.3 &51.5 \\
PSAC(R)        &70.4 &76.9 &55.7 \\
Heterogeneous Memory(HME)(R+C)~\cite{fan2019hetero} &73.9 &77.8 &53.8 \\
Location Aware GCN~\cite{huang2020locationaware} &74.3 &81.1 &56.3 \\
HCRN~\cite{minh2020hier}     &75.0 &81.4 &55.9 \\
Bridge2Answer~\cite{park2021bridge} & 75.9 &82.6 &57.5 \\
QueST~\cite{jiang2020divide}   &75.9 &81.0 &59.7 \\
ClipBERT~\cite{lei2021clipbert} 1x1 (Ntest=1) &82.9 &87.5 &59.4 \\
ClipBERT~\cite{lei2021clipbert} 1x1               &82.8 &87.8 &60.3 \\
\midrule 
Ours (Text-only, T5 pretrained~\cite{T5})             & 98.4 & 97.3 & 62.5 \\
Ours (Video+Text, Open Vocabulary)             & 11.8 & 12.5 & 27.3  \\
    \bottomrule
    \end{tabular}
    }
    \end{center}
\end{table*}

\section{Implementation and experimentation details}

\textbf{Model Training.} For pretraining, we train the model with a batch size of 256 for 500,000 steps. The learning rate was set to 0.001  When finetuning, the batch size was 128 and trained for 10,000 steps, with a learning rate of $1\times 10^{-6}$. We used the Adam optimizer, with weight decay set to 0.01. 

\subsection{Metrics}
For the IVQA dataset, we use accuracy as defined in the paper:
\begin{equation}
    Acc = \min \{ \cfrac{\# \text{humans with ans}}{2}, 1 \}
\end{equation}
averaged over 5 choose 4 of the ground truth (GT) answers.

For MSRVTT-QA and MSVD-QA, we use the standard accuracy metric. In the open-ended text generative setting, we check that the generated string is exactly equal to the ground truth string. The output string is the result of using beam search on the trained model. This is the hardest setting. With the restricted vocabulary, this check is easier, as there are fewer output options.

\section{Additional visualizations}

\begin{figure}[t]
   \centering
   \includegraphics[width=\linewidth]{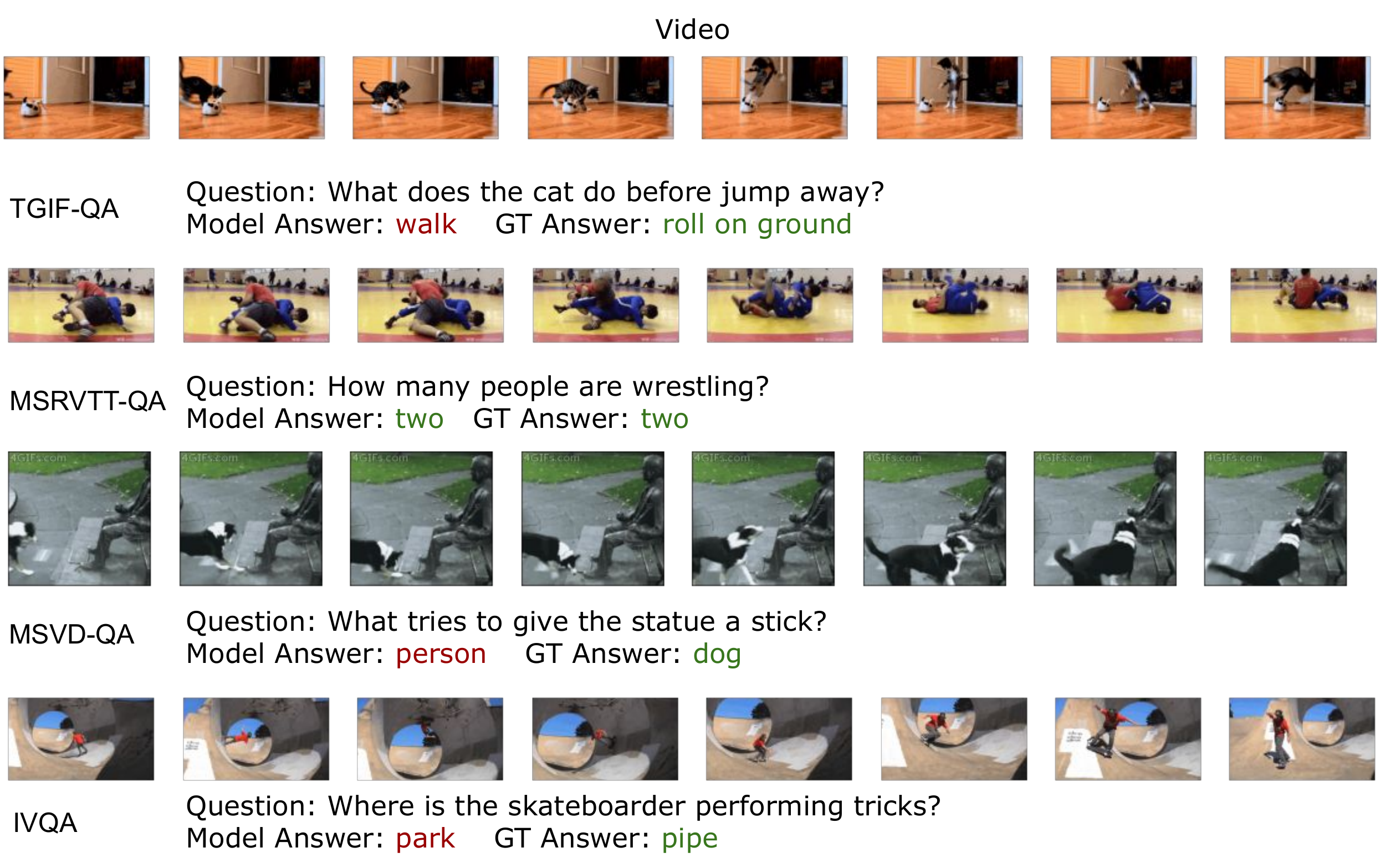}
  \caption{Examples of our method, from TGIF-QA, MSRVTT, MSVD, and IVQA datasets, in this order.  In some, the model gets the answer wrong, though it is reasonable, e.g., park vs pipe, while in others it is wrong, e.g., person and dog. In the TGIF-QA one, it is unclear if walk or roll on ground is correct.}
   \label{fig:examples}
 \end{figure}

Figure~\ref{fig:examples} shows additional visualizations of our approach.


\end{document}